\documentclass{article}

\usepackage{nips13submit_e,times}

\usepackage{geometry}                % See geometry.pdf to learn the layout options. There are lots.
\usepackage{graphicx}
\graphicspath{{figures/}}
\usepackage{amssymb}
\usepackage{amsmath}
\usepackage{epstopdf}
\usepackage{wrapfig}
\usepackage{ifthen}
\newcommand{\beq}[1][]{\begin{equation} \ifthenelse{\equal{#1}{}}{}{\label{#1}}}
\newcommand{\eeq}{\end{equation} }
\newcommand{\beqr}{\begin{eqnarray}}
\newcommand{\eeqr}{\end{eqnarray} }

 \newcommand{\req}[1]{(\ref{#1})}

\newcommand{\wa}{{W^{21}}}
\newcommand{\wb}{{W^{32}}}
\newcommand{\sio}{{\Sigma^{31}}}
\newcommand{\si}{\Sigma^{11}}
\newcommand{\rs}{{V^{11}}}
\newcommand{\ls}{{U^{33}}}
\newcommand{\sv}{{S^{31}}}

\newcommand{\wao}{{\overline W^{21}}}
\newcommand{\wbo}{{\overline W^{32}}}
\newcommand{\ddt}{\frac{d}{dt}}

\nipsfinalcopy

\title{Exact solutions to the nonlinear dynamics of learning in deep linear neural networks}

\author{{\large \bf Andrew M. Saxe (asaxe@stanford.edu)} \\
  Department of Electrical Engineering
  \vspace{-0.1in}
    \AND {\large \bf James L. McClelland (mcclelland@stanford.edu)} \\
    Department of Psychology
    \vspace{-0.1in}
    \AND {\large \bf Surya Ganguli (sganguli@stanford.edu)} \\
Department of Applied Physics \\
Stanford University, Stanford, CA 94305 USA}

\begin{document}
\maketitle
\vspace{-0.2in}
\begin{abstract}
Despite the widespread practical success of deep learning methods, our theoretical understanding of the dynamics of learning in deep neural networks remains quite sparse.  We attempt to bridge the gap between the theory and practice of deep learning by systematically analyzing learning dynamics for the restricted case of deep linear neural networks.  Despite the linearity of their input-output map, such networks have nonlinear gradient descent dynamics on weights that change with the addition of each new hidden layer. We show that deep linear networks exhibit nonlinear learning phenomena similar to those seen in simulations of nonlinear networks, including long plateaus followed by rapid transitions to lower error solutions, and faster convergence from greedy unsupervised pretraining initial conditions than from random initial conditions. We provide an analytical description of these phenomena by finding new exact solutions to the nonlinear dynamics of deep learning.  Our theoretical analysis also reveals the surprising finding that as the depth of a network approaches infinity, learning speed can nevertheless remain finite: for a special class of initial conditions on the weights, very deep networks incur only a finite, depth independent, delay in learning speed relative to shallow networks.  We show that, under certain conditions on the training data, unsupervised pretraining can find this special class of initial conditions, while scaled random Gaussian initializations cannot.  We further exhibit a new class of random orthogonal initial conditions on weights that, like unsupervised pre-training, enjoys depth independent learning times. We further show that these initial conditions also lead to faithful propagation of gradients even in deep nonlinear networks, as long as they operate in a special regime known as the edge of chaos.   

\end{abstract}

Deep learning methods have realized impressive performance in a range of applications, from visual object classification \cite{Krizhevsky,Le2012,Ciresan2012} to speech recognition \cite{Mohamed2012} and natural language processing \cite{Collobert2008,Socher2013}. These successes have been achieved despite the noted difficulty of training such deep architectures \cite{Hochreiter1991,Bengio1994,LeCun1998,Bottou2007,Erhan2009}.  Indeed, many explanations for the difficulty of deep learning have been advanced in the literature, including the presence of many local minima, low curvature regions due to saturating nonlinearities, and exponential growth or decay of back-propagated gradients \cite{Bengio2009,Glorot2010,Erhan2010,Dauphin2013}.  Furthermore, many neural network simulations have observed strikingly nonlinear learning dynamics, including long plateaus of little apparent improvement followed by almost stage-like transitions to better performance.  However, a quantitative, analytical understanding of the rich dynamics of deep learning remains elusive.  For example, what determines the time scales over which deep learning unfolds? How does training speed retard with depth? Under what conditions will greedy unsupervised pretraining speed up learning? And how do the final learned internal representations depend on the statistical regularities inherent in the training data? 

Here we provide an exact analytical theory of learning in deep linear neural networks that quantitatively answers these questions for this restricted setting. Because of its linearity, the input-output map of a deep linear network can always be rewritten as a shallow network. In this sense, a linear network does not gain expressive power from depth, and hence will underfit and perform poorly on complex real world problems. But while it lacks this important aspect of practical deep learning systems, a deep linear network can nonetheless exhibit highly nonlinear learning dynamics, and these dynamics change with increasing depth. Indeed, the training error, as a function of the network weights, is non-convex, and gradient descent dynamics on this non-convex error surface exhibits a subtle interplay between different weights across multiple layers of the network.  Hence deep linear networks provide an important starting point for understanding deep learning dynamics.

To answer these questions, we derive and analyze a set of nonlinear coupled differential equations describing learning dynamics on weight space as a function of the statistical structure of the inputs and outputs.   We find exact time-dependent solutions to these nonlinear equations, as well as find conserved quantities in the weight dynamics arising from symmetries in the error function. These solutions provide intuition into how a deep network successively builds up information about the statistical structure of the training data and embeds this information into its weights and internal representations.  Moreover, we compare our analytical solutions of learning dynamics in deep linear networks to numerical simulations of learning dynamics in deep non-linear networks, and find that our analytical solutions provide a reasonable approximation.  Our solutions also reflect nonlinear phenomena seen in simulations, including alternating plateaus and sharp periods of rapid improvement.  Indeed, it has been shown previously \cite{Anonymous.2013} that this nonlinear learning dynamics in deep linear networks is sufficient to qualitatively capture aspects of the progressive, hierarchical differentiation of conceptual structure seen in infant development.  Next, we apply these solutions to investigate the commonly used greedy layer-wise pretraining strategy for training deep networks \cite{Hinton2006a,Bengio}, and recover conditions under which such pretraining speeds learning. We show that these conditions are approximately satisfied for the MNIST dataset, and that unsupervised pretraining therefore confers an optimization advantage for deep linear networks applied to MNIST. Finally, we exhibit a new class of random orthogonal initial conditions on weights that, in linear networks, provide depth independent learning times, and we show that these initial conditions also lead to faithful propagation of gradients even in deep nonlinear networks. We further show that these initial conditions also lead to faithful propagation of gradients even in deep nonlinear networks, as long as they operate in a special regime known as the edge of chaos.  In this regime, synaptic gains are tuned so that linear amplification due to propagation of neural activity through weight matrices exactly balances dampening of activity due to saturating nonlinearities. In particular, we show that even in nonlinear networks, operating in this special regime, Jacobians that are involved in backpropagating error signals act like near isometries.

\section{General learning dynamics of gradient descent}
\label{learn_dyn}
\begin{wrapfigure}{r}{2.5in}
\vspace{-1.1in}
\begin{center}
\includegraphics[width=2.0in]{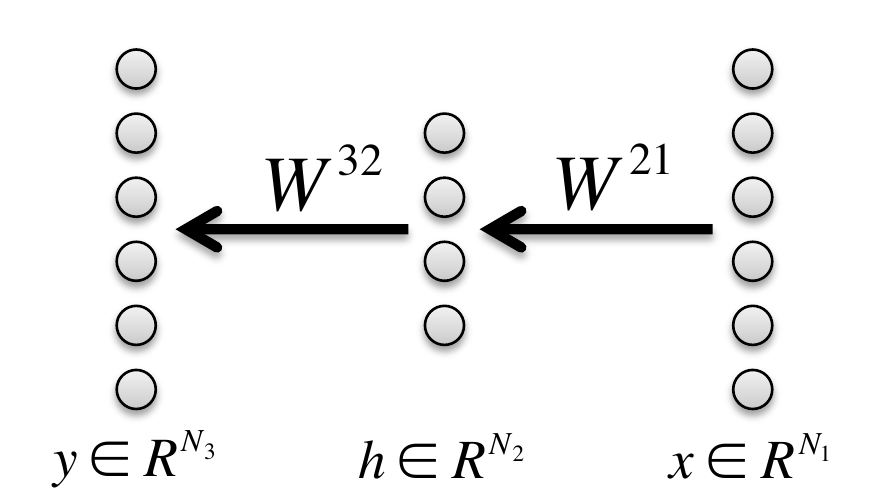}
\end{center}
\caption{The three layer network analyzed in this section.} 
\vspace{-.3in}
\label{network_fig}
\end{wrapfigure}

We begin by analyzing learning in a three layer network (input, hidden, and output) with linear activation functions (Fig \ref{network_fig}).  We let $N_i$ be the number of neurons in layer $i$.  The input-output map of the network is $y=\wb \wa x$.  
We wish to train the network to learn a particular input-output map from a set of $P$ training examples $\left\{x^\mu,y^\mu\right\}, \mu=1,\ldots, P$.  Training is accomplished via gradient descent on the squared error $\sum_{\mu=1}^P\left\| y^\mu - \wb \wa x^\mu\right\|^2$ between the desired feature output, and the network's feature output. This gradient descent procedure yields the batch learning rule
\beq
	\Delta \wa  =  \lambda \sum_{\mu=1}^{P}\wb^T \left( y^\mu x^{\mu T} - \wb \wa x^\mu x^{\mu T} \right), \qquad 
	\Delta \wb  =  \lambda \sum_{\mu=1}^{P} \left( y^\mu x^{\mu T} - \wb \wa x^\mu x^{\mu T} \right) \wa^T, \label{wb_delta}
\eeq
where $\lambda$ is a small learning rate. As long as $\lambda$ is sufficiently small, we can take a continuous time limit to obtain the dynamics,
\beq
	\tau \ddt \wa  =  \wb^T \left( \sio - \wb \wa \si \right), \qquad 
	\tau \ddt \wb  =    \left( \sio - \wb \wa \si \right) \wa^T,\label{wb_avg}
\eeq
where  $ \si \equiv  \sum_{\mu=1}^P x^\mu x^{\mu T}$ is an $N_1 \times N_1$ input correlation matrix, $\sio \equiv  \sum_{\mu=1}^P y^\mu x^{\mu T}$  is an $N_3 \times N_1$ input-output correlation matrix, and $\tau \equiv \frac{1}{\lambda}.$
Here $t$ measures time in units of iterations; as $t$ varies from 0 to 1, the network has seen $P$ examples corresponding to one iteration. Despite the linearity of the network's input-output map, the gradient descent learning dynamics given in Eqn \req{wb_avg} constitutes a complex set of coupled nonlinear differential equations with up to cubic interactions in the weights.

\subsection{Learning dynamics with orthogonal inputs}
\label{learn_dyn_orthog_sec}
Our fundamental goal is to understand the dynamics of learning in (\ref{wb_avg}) as a function of the input statistics $\si$ and input-output statistics $\sio$. In general, the outcome of learning will reflect an interplay between input correlations, described by $\si$, and the input-output correlations described by $\sio$. To begin, though, we further simplify the analysis by focusing on the case of orthogonal input representations where $\si=I$. This assumption will hold exactly for whitened input data, a widely used preprocessing step. %To allow for orthogonal input representations, we must have at least as many input units than examples, that is $N_1 \geq P$, and for simplicity we choose $N_1 = P$. 
%We further focus on the case in which $N_3 > N_1$, so that there are more output units than input units.

Because we have assumed orthogonal input representations ($\si = I$), the input-output correlation matrix contains all of the information about the dataset used in learning, and it plays a pivotal role in the learning dynamics. We consider its singular value decomposition (SVD)
\beq
	\sio = \ls \sv \rs^T = \textstyle \sum_{\alpha=1}^{N_1} s_\alpha u_\alpha v_\alpha^T, \label{eq_svd}
\eeq
which will be central in our analysis.  Here $\rs$ is an $N_1 \times N_1$ orthogonal matrix whose columns contain \textit{input-analyzing} singular vectors $v_\alpha$ that reflect independent modes of variation in the input, $\ls$ is an $N_3 \times N_3$ orthogonal matrix whose columns  contain \textit{output-analyzing} singular vectors $u_\alpha$ that reflect independent modes of variation in the output, and $\sv$ is an $N_3 \times N_1$ matrix whose only nonzero elements are on the diagonal; these elements are the singular values $s_\alpha,\alpha=1,\ldots,N_1$ ordered so that $s_1 \geq s_2 \geq \cdots\geq s_{N_1}$. 

Now, performing the change of variables on synaptic weight space, $\wa =  \wao \rs^T$,  $\wb = \ls \wbo$, the dynamics in \eqref{wb_avg} simplify to 
\beq
	\tau \ddt \wao =  \wbo^T(\sv - \wbo \wao), \qquad
	\tau \ddt \wbo =  (\sv - \wbo \wao) \wao^T. \label{wbo_dyn}
\eeq
To gain intuition for these equations, note that while the matrix elements of $\wa$ and $\wb$ connected neurons in one layer to neurons in the next layer, we can think of the matrix element $\wao_{i\alpha}$ as connecting input mode $v_\alpha$ to hidden neuron $i$, and the matrix element $\wbo_{\alpha i}$ as connecting hidden neuron $i$ to output mode $u_\alpha$. Let $a^\alpha$ be the $\alpha^{\textrm{th}}$ column of $\wao$, and let $b^{\alpha T}$ be the $\alpha^{\textrm{th}}$ row of $\wbo$. Intuitively, $a^\alpha$ is a column vector of $N_2$ synaptic weights presynaptic to the hidden layer coming from input mode $\alpha$, and $b^\alpha$ is a column vector of $N_2$ synaptic weights postsynaptic to the hidden layer going to output mode $\alpha$. In terms of these variables, or connectivity modes, the learning dynamics in (\ref{wbo_dyn}) become
\beq
	\tau \ddt a^\alpha  =  (s_\alpha - a^\alpha \cdot b^\alpha) \, b^\alpha - \sum_{\gamma \neq \alpha} b^\gamma \, (a^\alpha \cdot b^\gamma), \label{a_dyn} \qquad
	\tau \ddt b^\alpha  = ( s_\alpha - a^\alpha \cdot b^\alpha) \, a^\alpha - \sum_{\gamma \neq \alpha} a^\gamma \, (b^\alpha \cdot a^\gamma). %\label{b_dyn} 
\eeq
Note that $s_\alpha = 0$ for $\alpha > N_1$. These dynamics arise from gradient descent on the energy function
\beq
E = \frac{1}{2\tau} \sum_\alpha  (s_\alpha - a^\alpha \cdot b^\alpha)^2 + \frac{1}{2\tau} \sum_{\alpha \neq \beta} (a^\alpha \cdot b^\beta)^2 \label{ab_en},
\eeq
and display an interesting combination of cooperative and competitive interactions. Consider the first terms in each equation. In these terms, the connectivity modes from the two layers, $a^\alpha$ and $b^\alpha$ associated with the same input-output mode of strength $s^\alpha$,  cooperate with each other to drive each other to larger magnitudes as well as point in similar directions in the space of hidden units;  in this fashion these terms drive the product of connectivity modes $a^\alpha\cdot b^\alpha$ to reflect the input-output mode strength $s^\alpha$.  The second terms describe competition between the connectivity modes in the first ($a^\alpha$) and second ($b^\beta$) layers associated with different input modes $\alpha$ and $\beta$.  This yields a symmetric, pairwise repulsive force between all distinct pairs of first and second layer connectivity modes, driving the network to a decoupled regime in which the different connectivity modes become orthogonal.

\subsection{The final outcome of learning}

The fixed point structure of gradient descent learning in linear networks was worked out in \cite{Baldi1989}.  In the language of the connectivity modes, a necessary condition for a fixed point is $a^\alpha \cdot b^\beta = s_\alpha \delta_{\alpha\beta}$, 
while $a^\alpha$ and $b^\alpha$ are zero whenever $s_\alpha=0$. To satisfy these relations for undercomplete hidden layers ($N_2 < N_1, N_2 < N_3$), $a^\alpha$ and $b^\alpha$ can be nonzero for at most $N_2$ values of $\alpha$. Since there are $\textrm{rank}(\sio)\equiv r$ nonzero values of $s_\alpha$, there are $\begin{pmatrix} r \\ N_2\end{pmatrix}$ families of fixed points. However, all of these fixed points are unstable, except for the one in which only the first $N_2$ strongest modes, i.e. $a^\alpha$ and $b^\alpha$ for $\alpha=1,\ldots,N_2$ are active.
Thus remarkably, the dynamics in \eqref{a_dyn} has only saddle points and no non-global local minima \cite{Baldi1989}.   In terms of the original synaptic variables $\wa$ and $\wb$, all globally stable fixed points satisfy 
\beq
	\wb \wa = \textstyle \sum_{\alpha=1}^{N_2} s_\alpha u_\alpha v_\alpha^T.
\eeq
\begin{wrapfigure}{r}{2in}
\vspace{-.5in}
\begin{center}
\includegraphics[width=2in]{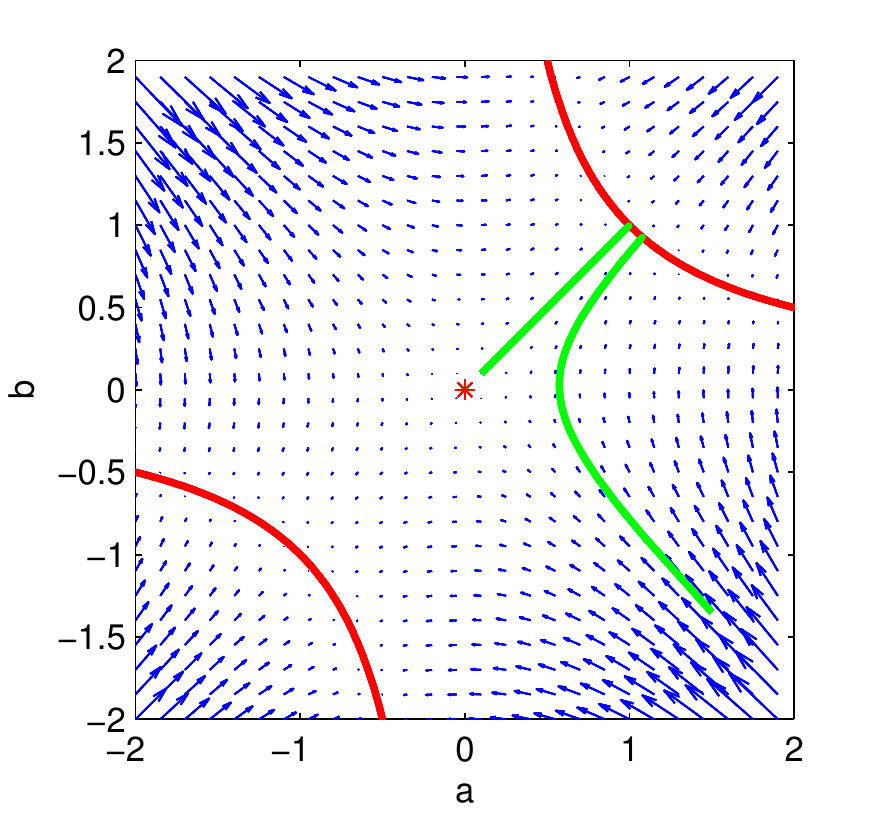}
\end{center}
\vspace{-.1in}
\caption{Vector field (blue), stable manifold (red) and two solution trajectories (green) for the two dimensional dynamics of $a$ and $b$ in \eqref{ab_dyn}, with $\tau=1,s=1$.}
\vspace{-.6362in}
\label{vector_field}
\end{wrapfigure}

Hence when learning has converged, the network will represent the closest rank $N_2$ approximation to the true input-output correlation matrix.  In this work, we are interested in understanding the dynamical weight trajectories and learning time scales that lead to this final fixed point.

\subsection{The time course of learning}

It is difficult though to exactly solve (\ref{a_dyn}) starting from arbitrary initial conditions because of the competitive interactions between different input-output modes. Therefore, to gain intuition for the general dynamics, we restrict our attention to a special class of initial conditions of the form $a^\alpha$ and $b^\alpha \propto r^\alpha$ for $\alpha=1,\ldots,N_2$, where $r^\alpha \cdot r^\beta = \delta_{\alpha \beta}$,
with all other connectivity modes $a^\alpha$ and $b^\alpha$ set to zero (see \cite{Fukumizu1998} for solutions to a partially overlapping but distinct set of initial conditions, further discussed in Supplementary Appendix \ref{hyper_dyn}). 
\begin{figure}[h]
\vspace{-.5in}
%\begin{wrapfigure}{l}{0.5\textwidth}
  \begin{minipage}[c]{0.35\textwidth}
    \includegraphics[width=\textwidth]{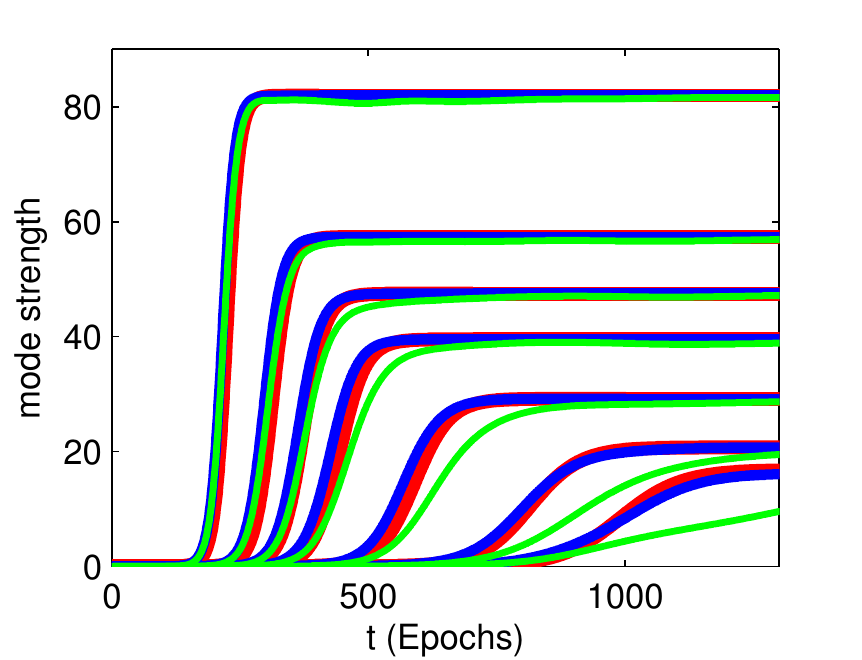}
  \end{minipage}\hfill
  \begin{minipage}[c]{0.6\textwidth}
    \includegraphics[width=\textwidth]{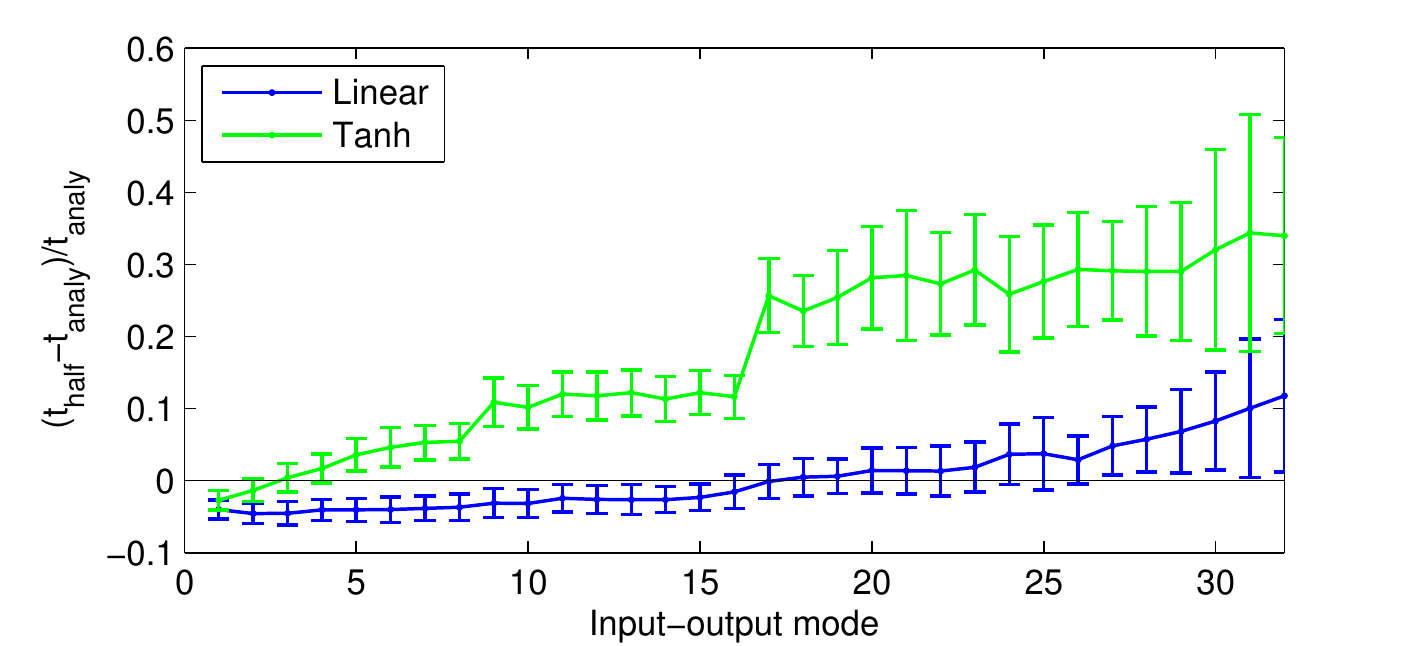}
  \end{minipage}\hfill
    \caption{
       \textbf{Left:} Dynamics of learning in a three layer neural network. Curves show the strength of the network's representation of seven modes of the input-output correlation matrix over the course of learning. Red traces show analytical curves from Eqn.~\ref{u_soln}. Blue traces show simulation of full dynamics of a linear network (Eqn. \req{wb_avg}) from small random initial conditions. Green traces show simulation of a nonlinear three layer network with tanh activation functions. To generate mode strengths for the nonlinear network, we computed the nonlinear network's evolving input-output correlation matrix, and plotted the diagonal elements of $\ls^T \sio_{tanh} \rs$ over time.  The training set consists of 32 orthogonal input patterns, each associated with a 1000-dimensional feature vector generated by a hierarchical diffusion process described in \cite{Anonymous.2013} with a five level binary tree and flip probability of 0.1. Modes 1, 2, 3, 5, 12, 18, and 31 are plotted with the rest excluded for clarity. Network training parameters were $\lambda=0.5e^{-3}, N_2=32, u_0=1e^{-6}$. \textbf{Right:} Delay in learning due to competitive dynamics and sigmoidal nonlinearities. Vertical axis shows the difference between simulated time of half learning and the analytical time of half learning, as a fraction of the analytical time of half learning. Error bars show standard deviation from 100 simulations with random initializations.} 
\label{mode_streng_traj}
\vspace{-.2in}
%\end{wrapfigure}
\end{figure}
Here $r^\alpha$ is a fixed collection of $N_2$ vectors that form an orthonormal basis for synaptic connections from an input or output mode onto the set of hidden units. Thus for this set of initial conditions, $a^\alpha$ and $b^\alpha$ point in the same direction for each alpha and differ only in their scalar magnitudes, and are orthogonal to all other connectivity modes. Such an initialization can be obtained by computing the SVD of $\sio$ and taking $\wb=\ls D_a R^T, \wa=R D_b \rs^T$ where $D_a,D_b$ are diagonal, and $R$ is an arbitrary orthogonal matrix; however, as we show 

in subsequent experiments, the solutions we find are also excellent approximations to trajectories from small random initial conditions. It is straightforward to verify that starting from these initial conditions, $a^\alpha$ and $b^\alpha$ will remain parallel to $r^\alpha$ for all future time.  Furthermore, because the different active modes are orthogonal to each other, they do not compete, or even interact with each other (all dot products in the second terms of \eqref{a_dyn}-\eqref{ab_en} are $0$).  

Thus this class of conditions defines an invariant manifold in weight space where the modes evolve independently of each other.

If we let $a = a^\alpha \cdot r^\alpha$,  $b = b^\alpha \cdot r^\alpha$, and $s = s^\alpha$, then the dynamics of the scalar projections $(a,b)$ obeys, 
\beq
	\tau \ddt a  =  b \, (s - a b), \qquad \tau \ddt b  =  a\, (s - a b). \label{ab_dyn}
\eeq

Thus our ability to decouple the connectivity modes yields a dramatically simplified two dimensional nonlinear system.  These equations can by solved by noting that they arise from gradient descent on the error,
\beq
	E(a,b)= \textstyle \frac{1}{2\tau}(s-ab)^2. \label{ab_2en}
\eeq
 %\begin{wrapfigure}{l}{0.5\textwidth}
%    \includegraphics[width=0.5\textwidth]{mode_strength_trajectory_three_layer.pdf} \includegraphics[width=0.5\textwidth]{fractional_delay.pdf}
%    \caption{
 %      \textbf{Top:} Dynamics of learning in a three layer neural network. Curves show the strength of the network's representation of seven modes of the input-output correlation matrix over the course of learning. Red traces show analytical curves from Eqn.~\ref{u_soln}. Blue traces show simulation of full dynamics of a linear network (Eqns.~(\ref{wb_avg})) from small random initial conditions. Green traces show simulation of a nonlinear three layer network with tanh activation functions. The analytical results align well with the full linear dynamics, and predict the ordering and approximate time course for the nonlinear sigmoidal network.  The training set consists of 32 orthogonal input patterns, each associated with a 1000-dimensional feature vector generated by a hierarchical diffusion process with a five level binary tree and flip probability of 0.1. Modes 1, 2, 3, 5, 12, 18, and 31 are plotted with the rest excluded for clarity. Network training parameters were $\lambda=0.5e^{-3}, N_2=32, u_0=1e^{-6}$. \textbf{Bottom:} Delay in learning due to competitive dynamics and sigmoidal nonlinearities. Vertical axis shows the difference between simulated time of half learning and the analytical time of half learning, as a fraction of the analytical time-to-half-learning.} 
%\label{mode_streng_traj}
%\vspace{-.7in}
%\end{wrapfigure}
%\end{figure}
This implies that the product $ab$ monotonically approaches the fixed point $s$ from its initial value. Moreover, $E(a,b)$ satisfies a symmetry under the one parameter family of scaling transformations $a \rightarrow \lambda a$, $b \rightarrow \frac{b}{\lambda}$.  This symmetry implies, through Noether's theorem, the existence of a conserved quantity, namely $a^2-b^2$, which is a constant of motion.  Thus the dynamics simply follows hyperbolas of constant $a^2-b^2$ in the $(a,b)$ plane until it approaches the hyperbolic manifold of fixed points, $ab=s$. The origin $a=0,b=0$ is also a fixed point, but is unstable. Fig.~\ref{vector_field} shows a typical phase portrait for these dynamics.

As a measure of the timescale of learning, we are interested in how long it takes for $ab$ to approach $s$ from any given initial condition. The case of unequal $a$ and $b$ is treated in the Supplementary Appendix \ref{hyper_dyn} due to space constraints. Here we pursue an explicit solution with the assumption that $a=b$, a reasonable limit when starting with small random initial conditions.  We can then track the dynamics of $u\equiv ab$, which from (\ref{ab_dyn}) obeys
\beq
	\tau \ddt u =2u(s-u). \label{sigmoidal_dyn}
\eeq
This equation is separable and can be integrated to yield
\beq
	t = \tau \int_{u_0}^{u_f} \frac{du}{2u(s-u)}=\frac{\tau}{2s}\ln \frac{u_f(s-u_0)}{u_0(s-u_f)}. \label{u_int}
\eeq
Here $t$ is the time it takes for $u$ to travel from $u_0$ to $u_f$. If we assume a small initial condition $u_0=\epsilon$, and ask when $u_f$ is within $\epsilon$ of the fixed point $s$, i.e.~$u_f=s-\epsilon$, then the learning timescale in the limit $\epsilon\rightarrow 0$ is $t = \tau/s\ln\left(s/\epsilon\right)=O(\tau/s)$ (with a weak logarithmic dependence on the cutoff). This yields a key result: the timescale of learning of each input-output mode $\alpha$ of the correlation matrix $\sio$ is inversely proportional to the correlation strength $s_\alpha$ of the mode. Thus the stronger an input-output relationship, the quicker it is learned.

We can also find the entire time course of learning by inverting \req{u_int} to obtain
\beq
	u_f(t)=\frac{se^{2st/\tau}}{e^{2st/\tau}-1+s/u_0}. \label{u_soln}
\eeq
This time course describes the temporal evolution of the product of the magnitudes of all weights from an input mode (with correlation strength $s$) into the hidden layers, and from the hidden layers to the same output mode.
 If this product starts at a small value $u_0 < s$, then it displays a sigmoidal rise which asymptotes to $s$ as $t\rightarrow\infty$. This sigmoid can exhibit sharp transitions from a state of no learning to full learning.   This analytical sigmoid learning curve is shown in Fig. \ref{mode_streng_traj} to yield a reasonable approximation to learning curves in linear networks that start from random initial conditions that are not on the orthogonal, decoupled invariant manifold--and that therefore exhibit competitive dynamics between connectivity modes--as well as in nonlinear networks solving the same task. We note that though the nonlinear networks behaved similarly to the linear case for this particular task, this is likely to be problem dependent.
 
\section{Deeper multilayer dynamics}

\label{deeper_multilayer_dyn}
%
%
%\begin{figure}
%  \begin{minipage}[c]{0.4\textwidth}
%    \includegraphics[width=\textwidth]{analytical_multilayer_soln_shape.pdf}
%  \end{minipage}\hfill
%  \begin{minipage}[c]{0.6\textwidth}
%    \caption{Solutions of Eqn.~(\ref{deep_dyn}) for $l$ values between 1 and 30. The time axis has been shifted for each curve such that the zero point marks the point at which $u$ reaches 90\% of its asymptotic value. Colorbar indicates $l$. Parameters are $s=1, \tau=l$. \ 
%    } \label{soln_shape}
%  \end{minipage}
%\end{figure}

The network analyzed in Section \ref{learn_dyn} is the minimal example of a multilayer net, with just a single layer of hidden units. How does gradient descent act in much deeper networks? We make an initial attempt in this direction based on initial conditions that yield particularly simple gradient descent dynamics.

In a linear neural network with $N_l$ layers and hence $N_l-1$ weight matrices indexed by $W^l, l=1,\cdots,N_l-1$, the gradient descent dynamics can be written as
\beq
	\tau \ddt W^l = \left(\prod_{i=l+1}^{N_l-1} W^i\right)^T\left[\sio - \left(\prod_{i=1}^{N_l-1}W^i\right) \si \right]\left(\prod_{i=1}^{l-1}W^{i}\right)^T, \label{multilayer_dyn}
\eeq
where $\prod_{i=a}^b W^{i} = W^bW^{(b-1)}\cdots W^{(a-1)}W^{a}$ with the special case that $\prod_{i=a}^b W^{i} = I$, the identity, if $a > b$. 

To describe the initial conditions, we suppose that there are $N_l$ orthogonal matrices $R_l$ that diagonalize the starting weight matrices, that is, $R_{l+1}^TW_l(0)R_l = D_l$ for all $l$, with the special case that $R_1= \rs$ and $R_{N_l} = \ls$. This requirement essentially demands that the output singular vectors of layer $l$ be the input singular vectors of the next layer $l+1$, so that a change in mode strength at any layer propagates to the output without mixing into other modes. We note that this formulation does not restrict hidden layer size; each hidden layer can be of a different size, and may be undercomplete or overcomplete. Making the change of variables $W_l = R_{l+1} \overline W_l R_{l}^T$ along with the assumption that $\si = I$ leads to a set of decoupled connectivity modes that evolve independently of each other.   In analogy to the simplification occurring in the three layer network from \eqref{wb_avg} to \eqref{ab_dyn}, each connectivity mode in the $N_l$ layered network can be described by $N_l-1$ scalars $a^1,\dots,a^{N_l-1}$, whose dynamics obeys gradient descent on the energy function (the analog of \eqref{ab_2en}),
\beq
	E(a_1,\cdots,a_{N_l-1})=\frac{1}{2\tau}\left(s-\prod_{i=1}^{N_l-1} a_i\right)^2.
\eeq
This dynamics also has a set of conserved quantities $a_i^2 - a_j^2$ arising from the energetic symmetry w.r.t. the transformation $a_i \rightarrow \lambda a_i$, $a_j \rightarrow \frac{a_j}{\lambda}$, and hence can be solved exactly.  We focus on the invariant submanifold in which $a_i(t=0)=a_0$ for all $i$, and track the dynamics of $u=\textstyle \prod_{i=1}^{N_l-1}a_i$, the overall strength of this mode, which obeys  (i.e. the generalization of  \eqref{sigmoidal_dyn}),
\beqr
	\tau \ddt u = (N_l-1) u^{2-2/(N_l-1)}(s - u). \label{deep_dyn}
\eeqr
This can be integrated for any positive integer $N_l$, though the expression is complicated. Once the overall strength increases sufficiently, learning explodes rapidly.

Eqn. (\ref{deep_dyn}) lets us study the dynamics of learning as depth limits to infinity. In particular, as $N_l\rightarrow \infty$ we have the dynamics
\beq
	\tau \ddt u = N_l u^2(s-u) \label{inf_dyn}
\eeq
which can be integrated to obtain
\beq
	t = \frac{\tau}{N_l} \left[ \frac{1}{s^2} \log \left( \frac{u_f(u_0-s)}{u_0(u_f-s)} \right) +\frac{1}{su_0}-\frac{1}{su_f}  \right]. \label{inf_tc}
\eeq
Remarkably this implies that, for a fixed learning rate, the learning time as measured by the number of iterations required tends to zero as $N_l$ goes to infinity. This result depends on the continuous time formulation, however. Any implementation will operate in discrete time and must choose a finite learning rate that yields stable dynamics. An estimate of the optimal learning rate can be derived from the maximum eigenvalue of the Hessian over the region of interest. \begin{wrapfigure}{r}{0.7\textwidth}

\begin{center}
\vspace{-.1in}
 \includegraphics[width=.7\textwidth]{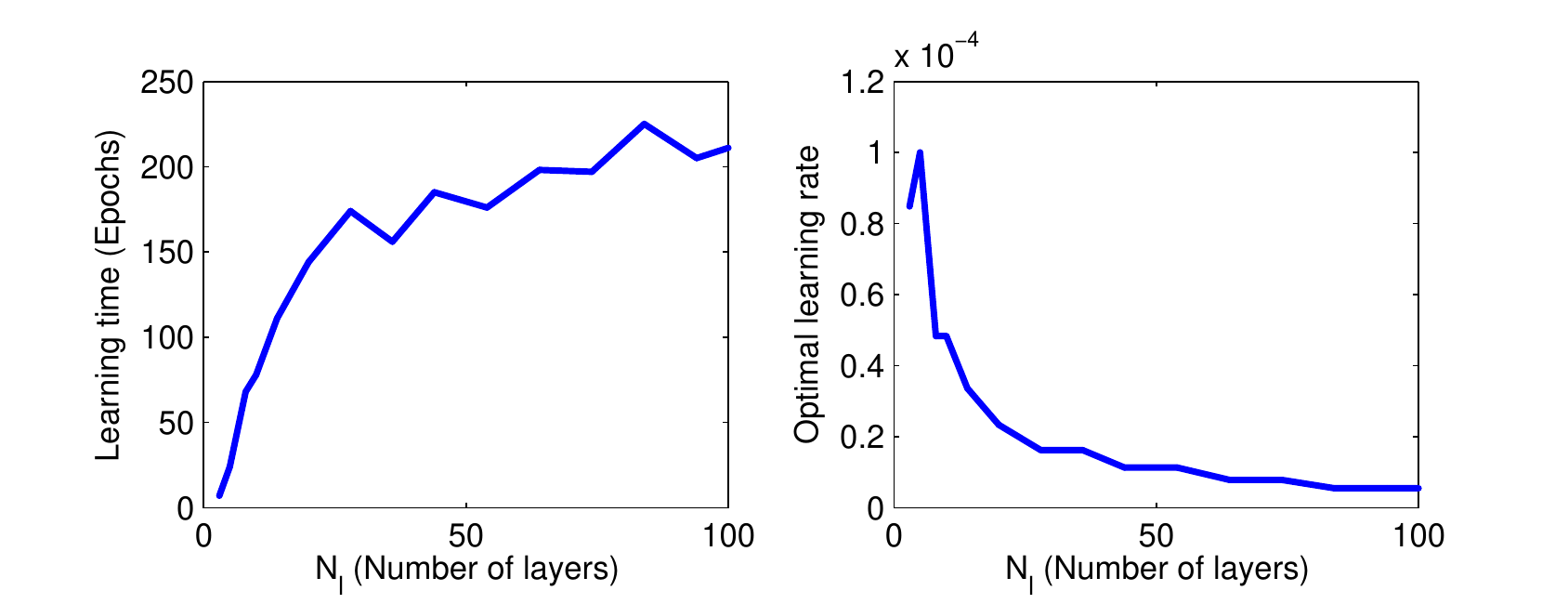}
\end{center}
\vspace{-.25in}
    \caption{
       \textbf{Left:} Learning time as a function of depth on MNIST. \textbf{Right:} Empirically optimal learning rates as a function of depth.} 
\label{depth_dependence_fig}
\vspace{-.3in}
\end{wrapfigure}
For linear networks with $a_i=a_j=a$, this optimal learning rate $\alpha_{opt}$ decays with depth as $ O\left(\frac{1}{N_ls^2}\right)$ for large $N_l$ (see Supplementary Appendix \ref{supp_opt_lr}).
 Incorporating this dependence of the learning rate on depth, the learning time as depth approaches infinity still surprisingly remains finite: with the optimal learning rate, the difference between learning times for an $N_l=3$ network and an $N_l=\infty$ network is $t_\infty-t_3 \sim O\left(s/\epsilon\right)$ for small $\epsilon$ (see Supplementary Appendix \ref{supp_opt_ls}). 
We emphasize that our analysis of learning speed is based on the number of iterations required, not the amount of computation--computing one iteration of a deep network will require more time than doing so in a shallow network.

To verify these predictions, we trained deep linear networks on the MNIST classification task with depths ranging from $N_l=3$ to $N_l=100$. We used hidden layers of size 1000, and calculated the iteration at which training error fell below a fixed threshold corresponding to nearly complete learning. We optimized the learning rate separately for each depth by training each network with twenty rates logarithmically spaced between $10^{-4}$ and $10^{-7}$ and picking the fastest. See Supplementary Appendix \ref{mnist_depth_expt} for full experimental details. Networks were initialized with decoupled initial conditions and starting initial mode strength $u_0=0.001$. Fig.~\ref{depth_dependence_fig} shows the resulting learning times, which saturate, and the empirically optimal learning rates, which scale like $O(1/N_l)$ as predicted.

Thus learning times in deep linear networks that start with decoupled initial conditions are only a finite amount slower than a shallow network regardless of depth. %Indeed, in cases where it is expensive to find the optimal learning rate empirically and a suboptimal learning rate is chosen, learning might even speed up in deep networks. 
Moreover, the delay incurred by depth scales inversely with the size of the initial strength of the association. Hence finding a way to initialize the mode strengths to large values is crucial for fast deep learning.

%The reason for this non-intuitive finding arises due to strong positive feedback associated with the cooperative interactions between the connectivity modes $a_i$ at each layer.  This positive feedback grows with the number of layers, and as long as the initial composite mode strength $u$ is above a threshold value, it rapidly amplifies to the final fixed point $s$ with a speed proportional to $N_l$. In contrast, if the initial mode strength decreases with increasing depth, then learning times can go to infinity. For instance, if each layer has a fixed strength $a_0$, then the overall starting mode strength $u_0=a_0^{N_l-1}$ exponentially decreases with depth for $a_0 < 1$, and this will send learning times in deep networks to infinity. Hence initializing the mode strengths to large values is crucial to fast learning.

\section{Finding good weight initializations: on greediness and randomness}

The previous subsection revealed the existence of a decoupled submanifold in weight space in which connectivity modes evolve independently of each other during learning, and learning times can be independent of depth, even for arbitrarily deep networks, as long as the initial composite, end to end mode strength, denoted by $u$ above, of every connectivity mode is $O(1)$. What numerical weight initilization procedures can get us close to this weight manifold, so that we can exploit its rapid learning properties?
\par A breakthrough in training deep neural networks started with the discovery that greedy layer-wise unsupervised pretraining could substantially speed up and improve the generalization performance of standard 
gradient descent \cite{Hinton2006a,Bengio}.  Unsupervised pretraining has been shown to speed the optimization of deep networks, and also to act as a special regularizer towards solutions with better generalization 
performance \cite{Bengio, Bengio2009, Glorot2010, Erhan2010}. At the same time, recent results have obtained excellent performance starting from carefully-scaled random initializations, though interestingly, pretrained initializations still exhibit faster convergence \cite{Martens2010,Glorot2010,Chapelle2011,Ciresan2012,Mohamed2012,Krizhevsky,Sutskever2013} (see Supplementary Appendix \ref{opt_advantage_lit_review} for discussion). Here we examine analytically how unsupervised pretraining achieves an optimization advantage, at least in deep linear networks, by finding the special class of orthogonalized, decoupled initial conditions in the previous section that allow for rapid supervised deep learning, for input-output tasks with a certain precise structure.  Subsequently, we analyze the properties of random initilizations.

We consider the following pretraining and finetuning procedure: First, using autoencoders as the unsupervised pretraining module \cite{Bengio, Bengio2009}, the network is trained to produce its input as its output ($y_{\textrm{pre}}^\mu=x^\mu$). Subsequently, the network is finetuned on the ultimate input-output task of interest (e.g., a classification task). In the following we consider the case $N_2=N_1$ for simplicity.

During the pretraining phase, the input-output correlation matrix $\sio_{\textrm{pre}}$ is simply the input correlation matrix $\si$. Hence the SVD of $\sio_{\textrm{pre}}$ is PCA on the input correlation matrix, since $\sio_{\textrm{pre}}=\si = Q\Lambda Q^T$, where $Q$ are eigenvectors of $\si$ and $\Lambda$ is a diagonal matrix of variances. Our analysis of the learning dynamics in Section \ref{learn_dyn_orthog_sec} does not directly apply, because here the input correlation matrix is not white. In Supplementary Appendix \ref{supp_learn_corr} we generalize our results to handle this case. During pretraining, the weights approach $\wb \wa = \sio(\sio)^{-1},$ but since they do not reach the fixed point in finite time, they will end at $\wb\wa=Q M Q^T$ where $M$ is a diagonal matrix that is approaching the identity matrix during learning. Hence in general, $\wb=QM^{1/2}C^{-1}$ and $\wa=CM^{1/2}Q^T$ where $C$ is any invertible matrix. When starting from small random weights, though, each weight matrix will end up with a roughly balanced contribution to the overall map. This corresponds to having $C\approx R_2$ where $R_2$ is orthogonal. Hence at the end of the pretraining phase, the input-to-hidden mapping will be $\wa = R_2M^{1/2}Q^T$ where $R_2$ is an arbitrary orthogonal matrix.

Now consider the fine-tuning phase. Here the weights are trained on the ultimate task of interest with input-output correlations $\sio=\ls \sv \rs$. The matrix $\wa$ begins from the pretrained initial condition $\wa = R_2M^{1/2}Q^T$. For the fine-tuning task, a decoupled initial condition for $\wa$ is one that can be written as $\wa=R_2D_1\rs^T$ (see Section \ref{deeper_multilayer_dyn}). Clearly, this will be possible only if
\beq
Q = \rs. \label{consist_cond}
\eeq

\begin{figure}
\vspace{-0.5in}
    \includegraphics[width=\textwidth]{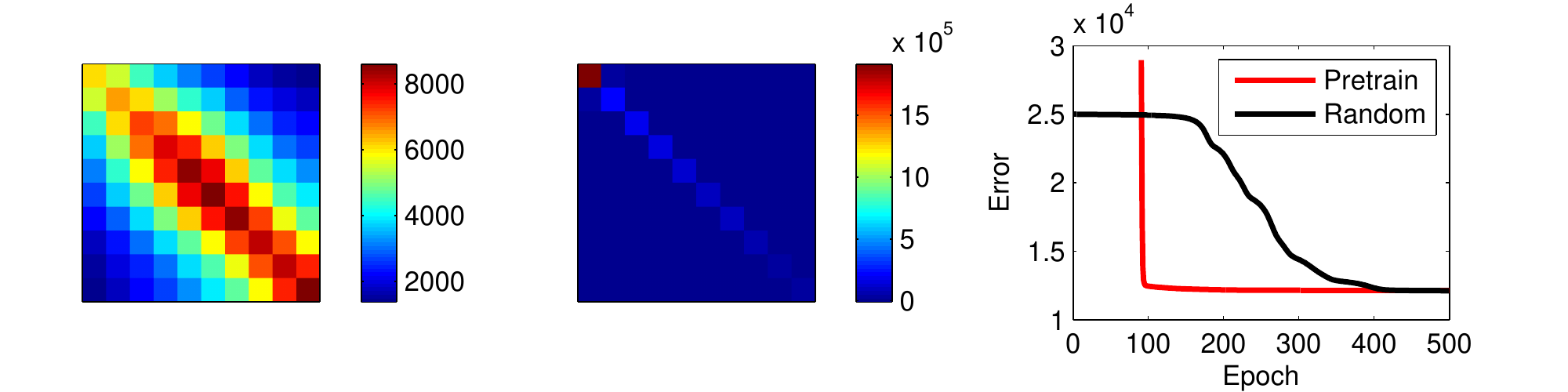}
	\vspace{-0.3in}
    \caption{
       MNIST satisfies the consistency condition for greedy pretraining. \textbf{Left:}  Submatrix from the raw MNIST input correlation matrix $\si$. \textbf{Center:} Submatrix of  $\rs \si \rs^T$ which is approximately diagonal as required. \textbf{Right:} Learning curves on MNIST for a five layer linear network starting from random (black) and pretrained (red) initial conditions. Pretrained curve starts with a delay due to pretraining time.  The small random initial conditions correspond to all weights chosen i.i.d. from a zero mean Gaussian with standard deviation 0.01.} 
\label{pretraining_fig}
\vspace{-.1in}
%\end{wrapfigure}
\end{figure}

Then the initial condition obtained from pretraining will also be a decoupled initial condition for the finetuning phase, with initial mode strengths $D_1=M^{1/2}$ near one. Hence we can state the underlying condition required for successful greedy pretraining in deep linear networks: the right singular vectors of the ultimate input-ouput task of interest $\rs$ must be similar to the principal components of the input data $Q$. This is a quantitatively precise instantiation of the intuitive idea that unsupervised pretraining can help in a subsequent supervised learning task if (and only if) the statistical structure of the input is consistent with the structure of input-output map to be learned.  Moreover, this quantitative instantiation of this intuitive idea gives a simple empirical criterion that can be evaluated on any new dataset: given the input-output correlation $\sio$ and input correlation $\si$, compute the right singular vectors $\rs$ of $\sio$ and check that $\rs \si \rs^T$ is approximately diagonal. If the condition in Eqn.~\req{consist_cond} holds, autoencoder pretraining will have properly set up decoupled initial conditions for $\wa$, with an appreciable initial association strength near $1$. This argument also goes through straightforwardly for layer-wise pretraining of deeper networks. Fig.~\ref{pretraining_fig} shows that this consistency condition empirically holds on MNIST, and that a pretrained deep linear neural network learns faster than one started from small random initial conditions, even accounting for pretraining time (see Supplementary Appendix \ref{mnist_pretrain_expt} for experimental details). We note that this analysis is unlikely to carry over completely to nonlinear networks. Some nonlinear networks are approximately linear (e.g., tanh nonlinearities) after initialization with small random initializations, and hence our solutions may describe these dynamics well early in learning. However as the network enters its nonlinear regime, our solutions should not be expected to remain accurate. 
\begin{figure}[h]
\vspace{-.1in}
%\begin{wrapfigure}{l}{0.5\textwidth}
    \includegraphics[width=\textwidth]{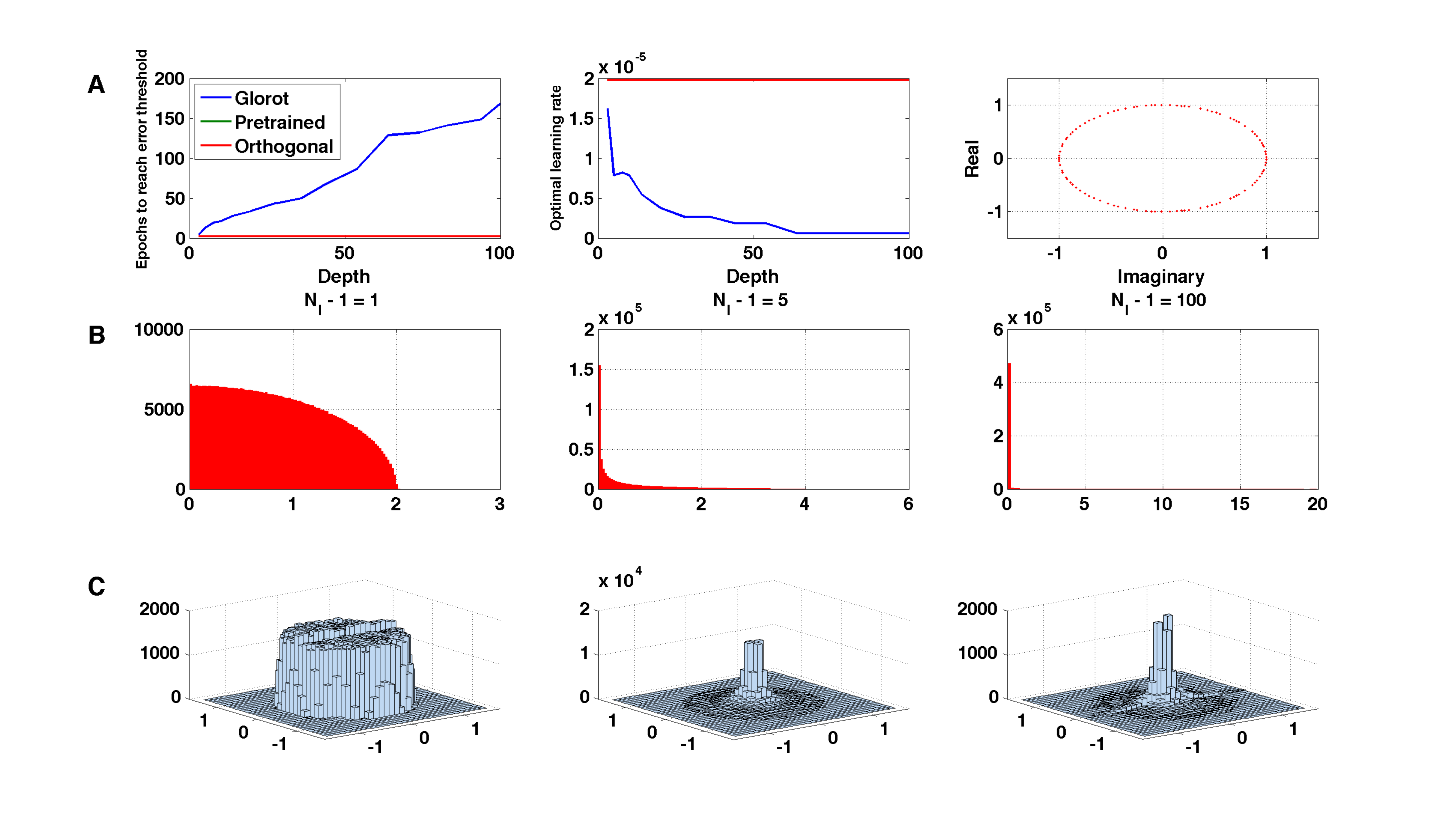}
    \vspace{-0.3in}
    \caption{
       \textbf{A} \textbf{Left:} Learning time (on MNIST using the same architecture and parameters as in Fig. \ref{depth_dependence_fig}) as a function of depth for different initial conditions on weights (scaled i.i.d. uniform weights chosen to preserve the norm of propagated gradients as proposed in \cite{Glorot2010} (blue), greedy unsupervised pre-training (green) and random orthogonal matrices (red).  The red curve lies on top of the green curve. \textbf{Middle:} Optimal learning rates as a function of depth for different weight initilizations. \textbf{Right:} The eigenvalue spectrum, in the complex plane, of a random $100$ by $100$ orthogonal matrix. \textbf{B} Histograms of the singular values of products of $N_l-1$ independent random Gaussian $N$ by $N$ matrices whose elements themselves are chosen i.i.d. from a zero mean Gaussian with standard deviation $1/\sqrt{N}$.  In all cases, $N=1000$, and histograms are taken over $500$ realizations of such random product matrices, yielding a total $5\cdot 10^5$ singular values in each histogram. \textbf{C} Histograms of the eigenvalue distributions on the complex plane of the same product matrices in \textbf{B}. The bin width is 0.1, and, for visualization purposes, the bin containing the origin has been removed in each case; this bin would otherwise dominate the histogram in the middle and right plots, as it contains $32\%$ and $94\%$ of the eigenvalues respectively. 
       }
\label{dynamic_crit}
\vspace{-.2in}
%\end{wrapfigure}
\end{figure}

As an alternative to greedy layerwise pre-training, \cite{Glorot2010} proposed choosing appropriately scaled initial conditions on weights that would preserve the norm of typical error vectors as they 
were backpropagated through the deep network. In our context, the appropriate norm-preserving scaling for the initial condition of an $N$ by $N$ connectivity matrix $W$ between any two layers corresponds to choosing each weight i.i.d. from a zero mean Gaussian with standard deviation $1/\sqrt{N}$.  With this choice, $\langle v^T W^T W v \rangle_W = v^T v$, where $\langle \cdot \rangle_W$ denotes an average over distribution of the random matrix $W$.  Moreover, the distribution of $v^T W^T W v$ concentrates about its mean for large $N$. Thus with this scaling, in linear networks, both the forward propagation of activity, and backpropagation of gradients is typically norm-preserving.  However, with this initialization, the learning time with depth on linear networks trained on MNIST grows with depth (Fig. \ref{dynamic_crit}A, left, blue). This growth is in distinct contradiction with the theoretical prediction, made above, of depth independent learning times starting from the decoupled submanifold of weights with composite mode strength $O(1)$.  This suggests that the scaled random initialization scheme, despite its norm-preserving nature, does not find this submanifold in weight space.  In contrast, learning times with greedy layerwise pre-training do not grow with depth (Fig. \ref{dynamic_crit}A, left, green curve hiding under red curve), consistent with the predictions of our theory (as a technical point: note that learning times under greedy pre-training initialization in Fig. \ref{dynamic_crit}A are faster than those obtained in  Fig. \ref{depth_dependence_fig} by explicitly choosing a point on the decoupled submanifold, because there the initial mode strength was chosen to be small ($u=0.001$) whereas greedy pre-training finds a composite mode strength closer to $1$). 

Is there a simple random initialization scheme that does enjoy the rapid learning properties of greedy-layerwise pre-training? We empirically show (Fig. \ref{dynamic_crit}A, left, red curve) that if we choose the initial weights in each layer to be a random orthogonal matrix (satisifying $W^T W = I$), instead of a scaled random Gaussian matrix, then this orthogonal random initialization condition yields depth independent learning times just like greedy layerwise pre-training (indeed the red and green curves are indistinguishable).  Theoretically, why do random orthogonal initializations yield depth independent learning times, but not scaled random Gaussian initializations, despite their norm preserving nature?

The answer lies in the eigenvalue and singular value spectra of products of Gaussian versus orthgonal random matrices. While a single random orthogonal matrix has eigenvalue spectra lying exactly on the unit circle in the complex plane (Fig. \ref{dynamic_crit}A right), the eigenvalue spectra of random Gaussian matrices, whose elements have variance $1/N$, form a uniform distribution on a solid disk of radius 1 the complex plane (Fig. \ref{dynamic_crit}C left).  Moreover the singular values of an orthogonal matrix are all exactly $1$, while the squared singular values of a scaled Gaussian random matrix have the well known Marcenko-Pasteur distribution, with a nontrivial spread even as $N \rightarrow \infty$,  (Fig. \ref{dynamic_crit}B left shows the distribution of singular values themselves).  Now consider a product of these matrices across all $N_l$ layers, representing the total end to end propagation of activity across a deep linear network:
\begin{equation}
W_{\text{Tot}} = \prod_{i=1}^{N_l-1} W^{(i+1,i)}.
\label{eq:Wtot}
\end{equation}
Due to the random choice of weights in each layer, $W_{\text{Tot}}$ is itself a random matrix. On average, it preserves the norm of a typical vector $v$ no matter whether the matrices in each layer are Gaussian or orthogonal.  However, the singular value spectra of $W_{\text{Tot}}$ differ markedly in the two cases.  Under random orthogonal initilization in each layer, $W_{\text{Tot}}$ is itself an orthogonal matrix and therefore has {\it all} singular values equal to $1$.  However, under random Gaussian initialization in each layer, there is as of yet no complete theoretical characterization of the singular value distribution of $W_{\text{Tot}}$. We have computed it numerically as a function of different depths in Fig. \ref{dynamic_crit}B, and we find that it develops a highly kurtotic nature as the depth increases. Most of the singular values become vanishingly small, while a long tail of very large singular values remain. Thus $W_{\text{Tot}}$ preserves the norm of a typical, randomly chosen vector $v$, but in a highly anisotropic manner, by strongly amplifying the projection of $v$ onto a very small subset of singular vectors and attenuating $v$ in all other directions.  Intuitively $W_{\text{Tot}}$, as well as the linear operator $W_{\text{Tot}}^T$ that would be closely related to backpropagation of gradients to early layers, act as amplifying projection operators at large depth $N_l$. In contrast, all of the eigenvalues of $W_{\text{Tot}}$ in the scaled Gaussian case concentrate closer to the origin as depth increases.  This discrepancy between the behavior of the eigenvalues and singular values of $W_{\text{Tot}}$, a phenomenon that could occur only if the eigenvectors of $W_{\text{Tot}}$ are highly non-orthogonal, reflects the highly non-normal nature of products of random Gaussian matrices (a non-normal matrix is by definition a matrix whose eigenvectors are non-orthogonal).   

While the combination of amplification and projection in $W_{\text{Tot}}$ can preserve norm, it is clear that it is not a good way to backpropagate errors; the projection of error vectors onto a high dimensional subspace corresponding to small singular values would be strongly attenuated, yielding vanishingly small gradient signals corresponding to these directions in the early layers. This effect, which is not present for random orthogonal initializations or greedy pretraining, would naturally explain the long learning times starting from scaled random Gaussian initial conditions relative to the other initilizations in Fig. \ref{dynamic_crit}A left. For both linear and nonlinear networks, a more likely appropriate condition on weights for generating fast learning times would be that of {\it dynamical isometry}. By this we mean that the product of Jacobians associated with error signal backpropagation should act as a near isometry, up to some overall global $O(1)$ scaling, on a subspace of as high a dimension as possible.  This is equivalent to having as many singular values of the product of Jacobians as possible within a small range around an $O(1)$ constant, and is closely related to the notion of restricted isometry in compressed sensing and random projections.  Preserving norms is a necessary but not sufficient condition for achieving dynamical isometry at large depths, as demonstrated in Fig. \ref{dynamic_crit}B, and we have shown that for linear networks, orthogonal initializations achieve exact dynamical isometry with all singular values at $1$, while greedy pre-training achieves it approximately.  

We note that the discrepancy in learning times between the scaled Gaussian initialization and the orthogonal or pre-training initializations is modest for the depths of around $6$ used in large scale applications, but is magnified at larger depths (Fig. \ref{dynamic_crit}A left). This may explain the modest improvement in learning times with greedy pre-training versus random scaled Gaussian initializations observed in applications (see discussion in Supplementary Appendix \ref{opt_advantage_lit_review}). We predict that this modest improvement will be magnified at higher depths, even in nonlinear networks.  Finally, we note that in recurrent networks, which can be thought of as infinitely deep feed-forward networks with tied weights, a very promising approach is a modification to the training objective that partially promotes dynamical isometry for the set of gradients currently being back-propagated \cite{pascanu2012difficulty}.

\section{Achieving approximate dynamical isometry in nonlinear networks}

We have shown above that deep random orthogonal linear networks achieve perfect dynamical isometry. Here we show that nonlinear versions of these networks can also achieve good dynamical isometry properties.  Consider the nonlinear feedforward dynamics 
\begin{equation}
x^{l+1}_i = \sum_j \, g \, W^{(l+1,l)}_{ij} \, \phi (x^l_j),
\label{eq:ffnldyn}
\end{equation}
where $x^l_i$ denotes the activity of neuron $i$ in layer $l$, $W^{(l+1,l)}_{ij}$ is a random orthogonal connectivity matrix from layer $l$ to $l+1$, $g$ is a scalar gain factor, and $\phi(x)$ is any nonlinearity that saturates as $x \rightarrow \pm \infty$.  We show in Supplementary appendix \ref{nonlin_orth_dyn} that there exists a critical value $g_c$ of the gain $g$ such that if $g < g_c$, activity will decay away to zero as it propagates through the layers, while if $g > g_c$, the strong linear positive gain will combat the damping due to the saturating nonlinearity, and activity will propagate indefinitely without decay, no matter how deep the network is.  When the nonlinearity is odd ($\phi(x) = -\phi(-x)$), so that the mean activity in each layer is approximately $0$, these dynamical properties can be quantitatively captured by the neural population variance in layer $l$,
\begin{equation}
q^{l} \equiv \frac{1}{N} \sum_{i=1}^N (x^l_i)^2.
\label{eq:ql}
\end{equation}

Thus $\lim_{l\rightarrow\infty} q^l \rightarrow 0$ for $g < g_c$ and  $\lim_{l\rightarrow\infty} q^l \rightarrow q^{\infty}(g) > 0$ for $g > g_c$.  When $\phi(x) = \tanh(x)$, we compute $g_c = 1$ and numerically compute $q^{\infty}(g)$ in Fig. \ref{nonlin_dyn} in Supplementary appendix \ref{nonlin_orth_dyn}. Thus these nonlinear feedforward networks exhibit a phase-transition at the critical gain; above the critical gain, infinitely deep networks exhibit chaotic percolating activity propagation, so we call the critical gain $g_c$ the edge of chaos, in analogy with terminology for recurrent networks. 
\begin{figure}[h]
\vspace{-.1in}
%\begin{wrapfigure}{l}{0.5\textwidth}
    \begin{center}
    \includegraphics[width=0.7\textwidth]{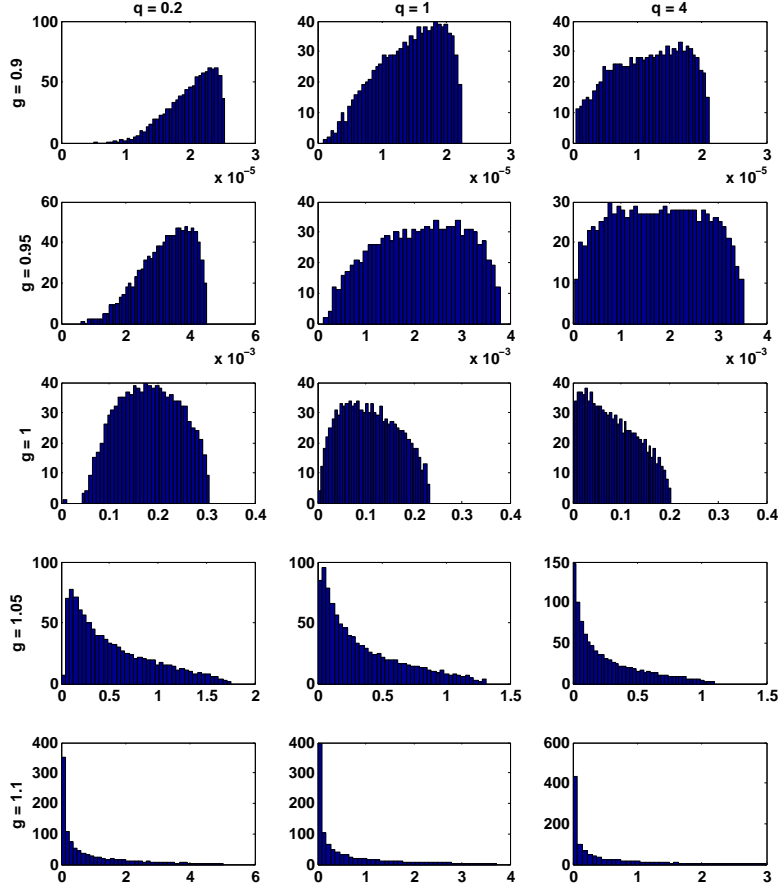}
    \end{center}
    \vspace{-0.4in}
    \caption{
       Singular value distribution of the end to end Jacobian, defined in \eqref{eq:eeJac}, for various values of the gain $g$ in \eqref{eq:ffnldyn} and the input layer population variance $q = q^1$ in \eqref{eq:ql}. The network architecture consists of $N_l=100$ layers with $N=1000$ neurons per layer, as in the linear case in Fig. \ref{dynamic_crit}B.}
\label{nonlin_sing}
\vspace{-0in}
%\end{wrapfigure}
\end{figure}

Now we are interested in how errors at the final layer $N_l$ backpropagate back to earlier layers, and whether or not these gradients explode or decay with depth.  To quantify this, for simplicity we consider the end to end Jacobian
\begin{equation}
J^{N_l, 1}_{ij}(x^{N_l}) \equiv \frac{\partial x^{N_l}_i}{\partial x^1_j}\bigg|_{x^{N_l}},
\label{eq:eeJac}
\end{equation}
which captures how input perturbations propagate to the output. If the singular value distribution of this Jacobian is well-behaved, with few extremely large or small singular values, then the backpropagation of gradients will also be well-behaved, and exhibit little explosion or decay. The Jacobian is evaluated at a particular point $x^{N_l}$ in the space of output layer activations, and this point is in turn obtained by iterating \eqref{eq:ffnldyn} starting from an initial input layer activation vector $x^{1}$.  Thus the singular value distribution of the Jacobian will depend not only on the gain $g$, but also on the initial condition $x^{1}$.  By rotational symmetry, we expect this distribution to depend on $x^{1}$, only through its population variance $q^1$.  Thus for large $N$, the singular value distribution of the end-to-end Jacobian in \eqref{eq:eeJac} (the analog of $W_{\text{Tot}}$ in \eqref{eq:Wtot} in the linear case), depends on only two parameters: gain $g$ and input population variance $q^1$.  

We have numerically computed this singular value distribution as a function of these two parameters in Fig. \ref{nonlin_sing}, for a single random orthogonal nonlinear network with $N=1000$ and $N_l=100$.  These results are typical; replotting the results for different random networks and different initial conditions (with the same input variance) yield very similar results.  We see that below the edge of chaos, when $g < 1$, the linear dampening over many layers yields extremely small singular values. Above the edge of chaos, when $g > 1$, the combination of positive linear amplification, and saturating nonlinear dampening yields an anisotropic distribution of singular values.  At the edge of chaos, $g=1$, an $O(1)$ fraction of the singular value distribution is concentrated in a range that remains $O(1)$ despite $100$ layers of propagation, reflecting appoximate dynamical isometry. Moreover, this nice property at $g=1$ remains valid even as the input variance $q^1$ is increased far beyond $1$, where the $\tanh$ function enters its nonlinear regime.  Thus the right column of Fig. \ref{nonlin_sing} at $g$ near $1$ indicates that the useful dynamical isometry properties of random orthogonal linear networks described above survives in nonlinear networks, even when activity patterns enter deeply into the nonlinear regime in the input layers.  Interestingly, the singular value spectrum is more robust to perturbations that increase $g$ from $1$ relative to those that decrease $g$. Indeed, the anisotropy in the singular value distribution at $g=1.1$ is relatively mild compared to that of random linear networks with scaled Gaussian initial conditions (compare the bottom row of Fig. \ref{nonlin_sing} with the right column of panel B in Fig. \ref{dynamic_crit}).  Thus overall, these numerical results suggest that being just beyond the edge of orthogonal chaos may be a good regime for learning in deep nonlinear networks.      

\vspace{-.1in}
\section{Discussion}

In summary, despite the simplicity of their input-output map, the dynamics of learning in deep linear networks reveals a surprising amount of rich mathematical structure, including nonlinear hyperbolic dynamics, plateaus and sudden performance transitions, a proliferation of saddle points, symmetries and conserved quantities, invariant submanifolds of independently evolving connectivity modes subserving rapid learning, and most importantly, a sensitive but computable dependence of learning time scales on input statistics, initial weight conditions, and network depth.  With the right initial conditions, deep linear networks can be only a finite amount slower than shallow networks, and unsupervised pretraining can find these initial conditions for tasks with the right structure. Moreover, we introduce a mathematical condition for faithful backpropagation of error signals, namely dynamical isometry, and show, surprisingly that random scaled Gaussian initializations cannot achieve this condition despite their norm-preserving nature, while greedy pre-training and random orthogonal initialization can, thereby achieving depth independent learning times.  Finally, we show that the property of dynamical isometry survives to good approximation even in extremely deep nonlinear random orthogonal networks operating just beyond the edge of chaos. At the cost of expressivity, deep linear networks gain theoretical tractability and may prove fertile for addressing other phenomena in deep learning, such as the impact of carefully-scaled initializations \cite{Glorot2010,Sutskever2013}, momentum \cite{Sutskever2013}, dropout regularization \cite{Krizhevsky}, and sparsity constraints \cite{Le2012}. While a full analytical treatment of learning in deep nonlinear networks currently remains open, one cannot reasonably hope to move towards such a theory without first completely understanding the linear case.  In this sense, our work fulfills an essential pre-requisite for progress towards a general, quantitative theory of deep learning.

\bibliographystyle{unsrt}
\bibliography{nips2013}

\section*{Supplementary Material}
\appendix
\section{Hyperbolic dynamics of learning}
\label{hyper_dyn}
In Section 1.3 of the main text we treat the dynamics of learning in three layer networks where mode strengths in each layer are equal, i.e, $a=b$, a reasonable limit when starting with small random initial conditions. More generally, though, we are interested in how long it takes for $ab$ to approach $s$ from any given initial condition. To access this, given the hyperbolic nature of the dynamics, it is useful to make the hyperbolic change of coordinates,

\beqr
	a=\sqrt{c_0} \cosh \frac{\theta}{2} \quad b=\sqrt{c_0}\sinh\frac{\theta}{2} \quad \textrm{for} ~~a^2>b^2 \\
	a=\sqrt{c_0} \sinh \frac{\theta}{2} \quad b=\sqrt{c_0}\cosh\frac{\theta}{2} \quad \textrm{for} ~~a^2<b^2.
\eeqr
Thus $\theta$ parametrizes the dynamically invariant manifolds $a^2-b^2=\pm c_0$. For any $c_0$ and $\theta$, this coordinate system covers the region $a+b>0$, which is the basin of attraction of the upper right component of the hyperbola $ab=s$. A symmetric situation exists for $a+b<0$, which is attracted to the lower left component of $ab=s$. We use $\theta$ as a coordinate to follow the dynamics of the product $ab$, and using the relations $ab=c_0 \sinh \theta$ and $a^2+b^2=c_0\cosh \theta$, we obtain
\beq
	\tau \frac{d\theta}{dt}=s-c_0 \sinh \theta.
\eeq
This differential equation is separable in $\theta$ and $t$ and can be integrated to yield
\beq
	t = \tau \int_{\theta_0}^{\theta_f} \frac{d\theta}{s-c_0\sinh\theta}=\frac{\tau}{\sqrt{c^2_0 + s^2}}\left[\ln \frac{ \sqrt{c^2_0 + s^2} + c_0 + s\tanh \frac{\theta}{2}}{\sqrt{c^2_0 + s^2} - c_0 - s\tanh \frac{\theta}{2} }\right]_{\theta_0}^{\theta_f}.
\eeq
Here $t$ is the amount of time it takes to travel from $\theta_0$ to $\theta_f$ along the hyperbola $a^2-b^2=\pm c_0$. The fixed point lies at $\theta = \sinh^{-1}s/c_0$, but the dynamics cannot reach the fixed point in finite time. Therefore we introduce a cutoff $\epsilon$ to mark the endpoint of learning, so that $\theta_f$ obeys $\sinh \theta_f=(1-\epsilon)s/c_0$ (i.e. $ab$ is close to $s$ by a factor $1-\epsilon$). We can then average over the initial conditions $c_0$ and $\theta_0$ to obtain the expected learning time of an input-output relation that has  a correlation strength $s$. Rather than doing this, it is easier to obtain a rough estimate of the timescale of learning
under the assumption that the initial weights are small, so that $c_0$ and $\theta_0$ are close to $0$. In this case $t=O(\tau/s)$ (with a weak logarithmic dependence on the cutoff (i.e. $\ln(1/\epsilon)$). This modestly generalizes the result given in the main text: the timescale of learning of each input-output mode $\alpha$ of the correlation matrix $\sio$ is inversely proportional to the correlation strength $s_\alpha$ of the mode even when $a$ and $b$ differ slightly, i.e., $c_0$ small. This is not an unreasonable limit for random initial conditions because $|c_0|=|a \cdot a - b \cdot b|$ where $a$ and $b$ are random vectors of $N_2$ synaptic weights into and out of the hidden units. Thus we expect the lengths of the two random vectors to be approximately equal and therefore $c_0$ will be small relative to the length of each vector.
\par These solutions are distinctly different from solutions for learning dynamics in three layer networks found in \cite{Fukumizu1998}.  In our notation, in \cite{Fukumizu1998}, it was shown that if the initial vectors $a^\alpha$ and $b^\alpha$ satisfy the matrix identity $\sum_\alpha a^\alpha a^{\alpha^T} = \sum_\alpha b^\alpha b^{\alpha^T}$ then the dynamics of learning becomes equivalent to a matrix Riccatti equation.  However, the hyperbolic dynamics derived here arises from a set of initial conditions that do not satisfy the restrictions of \cite{Fukumizu1998} and therefore do not arise through a solution to a matrix Ricatti equation.  Moreover, in going beyond a statement of the matrix Riccatti solution, our analysis provides intuition about the time-scales over which the learning dynamics unfolds, and crucially, our methods extend beyond the three layer case to the arbitrary $N_l$ layer case, not studied in \cite{Fukumizu1998}.

\section{ Optimal discrete time learning rates}
\label{supp_opt_lr}
In Section 2 we state results on the optimal learning rate as a function of depth in a deep linear network, which we derive here. Starting from the decoupled initial conditions given in the main text, the dynamics arise from gradient descent on
\beq
	E(a_1,\cdots,a_{N_l-1})=\frac{1}{2\tau}\left(s - \prod_{k=1}^{N_l-1}a_k\right).
\eeq
Hence for each $a_i$ we have 
\beq
	\frac{\partial E}{\partial a_i} = -\frac{1}{\tau}\left(s - \prod_{k=1}^{N_l-1}a_k\right)\left( \prod_{k\neq i}^{N_l-1}a_k\right) \equiv f(a_i)
\eeq
The elements of the Hessian are thus 
\beqr
	\frac{\partial^2 E}{\partial a_ia_j} & = & \frac{1}{\tau}\left( \prod_{k\neq j}^{N_l-1}a_k\right)\left( \prod_{k\neq i}^{N_l-1}a_k\right) - \frac{1}{\tau}\left(s - \prod_{k=1}^{N_l-1}a_k\right)\left( \prod_{k\neq i,j}^{N_l-1}a_k\right) \\
	& \equiv & g(a_i,a_j)
\eeqr 
for $i\neq j$, and
\beq
	\frac{\partial^2 E}{\partial a_i^2} = \frac{1}{\tau}\left( \prod_{k\neq i}^{N_l-1}a_k\right)^2 \equiv h(a_i)
\eeq
for $i = j$. 

We now assume that we start on the symmetric manifold, such that $a_i=a_j=a$ for all $i,j$. Thus we have

\beqr
	E(a)&=&\frac{1}{2\tau}\left(s - a^{N_l-1}\right),\\
	f(a)&= &-\frac{1}{\tau}\left(s - a^{N_l-1}\right)a^{N_l-2},\\
	g(a) &  =&  \frac{2}{\tau}a^{2N_l-4} - \frac{1}{\tau}sa^{N_l-3} \\
	h(a) &= &\frac{1}{\tau} a^{2N_l-4}
\eeqr

The Hessian is
\beq
	H(a) = \begin{bmatrix} 
		h & g & \cdots &g& g \\ 
		g & h  & \cdots & g&g \\
		\vdots & & \ddots && \vdots \\
		g & g &\cdots& h & g \\
		g & g& \cdots & g & h
			\end{bmatrix}.
\eeq
One eigenvector is $v_1=[1 1 \cdots 1]^T$ with eigenvalue $\lambda_1 = h + (N_l-2)g$, or
\beqr
	\lambda_1 = (2N_l-3)\frac{1}{\tau}a^{2N_l-4} - (N_l-2)\frac{1}{\tau}sa^{N_l-3}. \label{lambda}
\eeqr
Now consider the second order update (Newton-Raphson) (here we use $1$ to denote a vector of ones)
\beqr 
	a^{t+1}1 & = & a^t1 - H^{-1}f(a^t)1 \\
		& = & a^t 1 - f(a^t)H^{-1}1 \\
		 a^{t+1} & = & a^t - f(a^t)/\lambda_1(a^t)
\eeqr
Note that the basin of attraction does not include small initial conditions, because for small $a$ the Hessian is not positive definite.

To determine the optimal learning rate for first order gradient descent, we compute the maximum of $\lambda_1$ over the range of mode strengths that can be visited during learning, i.e., $a\in[0,s^{1/(N_l-1)}]$. This  occurs at the optimum, $a_{opt}=s^{1/(N_l-1)}$. Hence substituting this into \eqref{lambda} we have
\beq
	\lambda_1(a_{opt}) =  (N_l-1)\frac{1}{\tau}s^{\frac{2N_l-4}{N_l-1}} . \label{opt_lr}
\eeq	
The optimal learning rate $\alpha$ is proportional to $1/\lambda_1(a_{opt})$, and hence scales as
\beq
	\alpha \sim O\left(\frac{1}{N_ls^2}\right)
\eeq
for large $N_l$.

\subsection{Learning speeds with optimized learning rate}
\label{supp_opt_ls}
How does the optimal learning rate impact learning speeds? We compare the three layer learning time to the infinite depth limit learning time, with learning rate set inversely proportional to Eqn.~\req{opt_lr} with proportionality constant $c$. 

This yields a three layer learning time $t_3$ of
\beq
	t_3 = c \ln \frac{u_f(s-u_0)}{u_0(s-u_f)}
\eeq
and an infinite layer learning time $t_\infty$ of
\beq
	t_\infty = c \left[  \log \left( \frac{u_f(u_0-s)}{u_0(u_f-s)} \right) +\frac{s}{u_0}-\frac{s}{u_f}  \right],
\eeq

Hence the difference is
\beq
t_\infty-t_3 = \frac{cs}{u_0}-\frac{cs}{u_f} \approx \frac{cs}{\epsilon}
\eeq
where the final approximation is for $u_0=\epsilon,u_f=s-\epsilon$, and $\epsilon$ small. Thus very deep networks incur only a finite delay relative to shallow networks.

\section{Experimental setup for MNIST depth experiment}
\label{mnist_depth_expt}

We trained deep linear networks on the MNIST dataset with fifteen different depths  $N_l=\{3, 5, 8, 10, 14, 20, 28, 36, 44, 54, 64, 74, 84, 94, 100\}$. Given a 784-dimensional input example, the network tried to predict a 10-dimensional output vector containing a 1 in the index for the correct class, and zeros elsewhere. The network was trained using batch gradient descent via Eqn.~\req{multilayer_dyn} on the 50,000 sample MNIST training dataset. We note that Eqn.~\req{multilayer_dyn} makes use of the linearity of the network to speed training and reduce memory requirements. Instead of forward propagating all 50,000 training examples, we precompute $\sio$ and forward propagate only it. This enables experiments on very deep networks that otherwise would be computationally infeasible. Experiments were accelerated on GPU hardware using the GPUmat package. We used overcomplete hidden layers of size 1000. Here the overcompleteness is simply to demonstrate the applicability of the theory to this case; overcompleteness does not improve the representational power of the network. Networks were initialized with decoupled initial conditions and starting initial mode strength $u_0=0.001$, as described in the text. The random orthogonal matrices $R_l$ were selected by generating random Gaussian matrices and computing a QR decomposition to obtain an orthogonal matrix. Learning times were calculated as the iteration at which training error fell below a fixed threshold of $1.3\times10^4$ corresponding to nearly complete learning. Note that this level of performance is grossly inferior to what can be obtained using nonlinear networks, which reflects the limited capacity of a linear network. We optimized the learning rate $\lambda$ separately for each depth by training each network with twenty rates logarithmically spaced between $10^{-4}$ and $10^{-7}$ and picking the one that yielded the minimum learning time according to our threshold criterion. The range $10^{-4}$ and $10^{-7}$ was selected via preliminary experiments to ensure that the optimal learning rate always lay in the interior of the range for all depths. 

\section{Efficacy of unsupervised pretraining}
\label{opt_advantage_lit_review}

Recently high performance has been demonstrated in deep networks trained from random initial conditions \cite{Martens2010,Glorot2010,Chapelle2011,Ciresan2012,Mohamed2012,Krizhevsky,Sutskever2013}, suggesting that deep networks may not be as hard to train as previously thought. These results show that pretraining is not necessary to obtain state-of-the-art performance, and to achieve this they make use of a variety of techniques including carefully-scaled random initializations, more sophisticated second order or momentum-based optimization methods, and specialized convolutional architectures. It is therefore important to evaluate whether unsupervised pretraining is still useful, even if it is no longer necessary, for training deep networks. In particular, does pretraining still confer an optimization advantage and generalization advantage when used in conjunction with these new techniques? Here we review results from a variety of papers, which collectively show that unsupervised pretraining still confers an optimization advantage and a generalization advantage.

\subsection{Optimization advantage}
The optimization advatage of pretraining refers to faster convergence to the local optimum (i.e., faster learning speeds) when starting from pretrained initializations as compared to random initializations. Faster learning speeds starting from pretrained initial conditions have been consistently found with Hessian free optimization \cite{Martens2010,Chapelle2011}. This finding holds for two carefully-chosen random initialization schemes, the sparse connectivity scheme of \cite{Martens2010}, and the dense scaled scheme of \cite{Glorot2010} (as used by \cite{Chapelle2011}). Hence pretraining still confers a convergence speed advantage with second order methods. Pretrained initial conditions also result in faster convergence than carefully-chosen random initializations when optimizing with stochastic gradient descent \cite{Chapelle2011, Glorot2010}. In light of this, it appears that pretrained initial conditions confer an optimization advantage beyond what can be obtained currently with carefully-scaled random initializations, regardless of optimization technique. If run to convergence, second order methods and well-chosen scalings can erase the discrepancy between the final objective value obtained on the training set for pretrained relative to random initializations \cite{Martens2010,Chapelle2011}. The optimization advantage is thus purely one of convergence speed, not of finding a better local minimum. This coincides with the situation in linear networks, where all methods will eventually attain the same global minimum, but the rate of convergence can vary. Our analysis shows why this optimization advantage due to pretraining persists over well-chosen random initializations.

Finally, we note that Sutskever et al.~show that careful random initialization paired with carefully-tuned momentum can achieve excellent performance \cite{Sutskever2013}, but these experiments did not try pretrained initial conditions. Krizhevsky et al.~used convolutional architectures and did not attempt pretraining \cite{Krizhevsky}. Thus the possible utility of pretraining in combination with momentum, and in combination with convolutional architectures, dropout, and large supervised datasets, remains unclear. 

\subsection{Generalization advantage}
Pretraining can also act as a special regularizer, improving generalization error in certain instances. This generalization advantage appears to persist with new second order methods \cite{Martens2010,Chapelle2011}, and in comparison to gradient descent with careful random initializations \cite{Glorot2010,Chapelle2011,Lamblin2010,Mohamed2012}. An analysis of this effect in deep linear networks is out of the scope of this work, though promising tools have been developed for the three layer linear case \cite{Fukumizu1998}.

\section{Learning dynamics with task-aligned input correlations}
\label{supp_learn_corr}
In the main text we focused on orthogonal input correlations ($\si=I$) for simplicity, and to draw out the main intuitions. However our analysis can be extended to input correlations with a very particular structure. Recall that we decompose the input output correlations using the SVD as $\sio=\ls \sv \rs^T$. We can generalize our solutions to allow input correlations of the form $\si=\rs D\rs^T$. Intuitively, this condition requires the axes of variation in the input to coincide with the axes of variation in the input-output task, though the variances may differ. If we take $D=I$ then we recover the whitened case $\si=I$, and if we take $D=\Lambda$, then we can treat the autoencoding case. The final fixed points of the weights are given by the best rank $N_2$ approximation to $\sio(\si)^{-1}$. Making the same change of variables as in Eqn.~\eqref{wbo_dyn} we now obtain
\beq
	\tau \ddt \wao =  \wbo^T(\sv - \wbo \wao D), \qquad
	\tau \ddt \wbo =  (\sv - \wbo \wao D) \wao^T. \label{wbo_dyn_input_corr}
\eeq
which, again, is decoupled if $\wbo$ and $\wao$ begin diagonal. Based on this it is straightforward to generalize our results for the learning dynamics. 

\section{MNIST pretraining experiment}
\label{mnist_pretrain_expt}
We trained networks of depth 5 on the MNIST classification task with 200 hidden units per layer, starting either from small random initial conditions with each weight drawn independently from a Gaussian distribution with standard deviation 0.01, or from greedy layerwise pretrained initial conditions. For the pretrained network, each layer was trained to reconstruct the output of the next lower layer. In the finetuning stage, the network tried to predict a 10-dimensional output vector containing a 1 in the index for the correct class, and zeros elsewhere. The network was trained using batch gradient descent via Eqn.~\req{multilayer_dyn} on the 50,000 sample MNIST training dataset.  Since the network is linear, pretraining initializes the network with principal components of the input data, and, to the extent that the consistency condition of Eqn.~\req{consist_cond} holds, decouples these modes throughout the deep network, as described in the main text. 

\section{Analysis of Neural Dynamics in Nonlinear Orthogonal Networks}
\label{nonlin_orth_dyn}

We can derive a simple, analytical recursion relation for the propagation of neural population variance $q^l$, defined in \eqref{eq:ql}, across layers $l$ under the nonlinear dynamics \eqref{eq:ffnldyn}.  We have
\begin{equation}
q^{l+1} = \frac{1}{N} \sum_{i=1}^N (x^{l+1}_i)^2 = g^2 \frac{1}{N} \sum_{i=1}^N \phi(x^{l}_i)^2, 
\end{equation}
due to the dynamics in \eqref{eq:ffnldyn} and the orthogonality of $W^{(l+1,l)}$.  Now we know that by definition, the layer $l$ population $x^l_i$ has normalized variance $q^l$.  If we further assume that the distribution of activity across neurons in layer $l$ is well approximated by a Gaussian distribution, we can replace the sum over neurons $i$ with an integral over a zero mean unit variance Gaussian variable $z$:
\begin{equation}
q^{l+1} = g^2 \, \int \mathcal{D}z \,\phi \big(\sqrt{q^l} z \big)^2, 
\label{eq:qrec}
\end{equation}
where $ \mathcal{D}z \equiv \frac{1}{\sqrt{2 \pi}}e^{-\frac{1}{2} z^2} \, dz$ is the standard Gaussian measure.
\begin{figure}[h]
\vspace{-.1in}
%\begin{wrapfigure}{l}{0.5\textwidth}
    \begin{center}
    \includegraphics[width=0.7\textwidth]{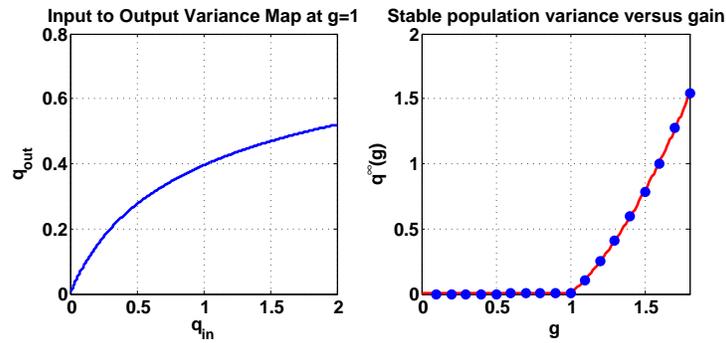}
    \end{center}
    \vspace{-0.3in}
    \caption{\textbf{Left}: The map from variance in the input layer $q^{in} = q^l$ to variance in the output layer $q^{out} = q^{l+1}$ in \eqref{eq:qrec} for $g=1$ and $\phi(x) = \tanh(x)$.  \textbf{Right}: The stable fixed points of this map, $q^{\infty}(g)$, as a function of the gain $g$. The red curve is the analytic theory obtained by numerically solving \eqref{eq:qfixed}. The blue points are obtained via numerical simulations of the dynamics in \eqref{eq:ffnldyn} for networks of depth $N_l = 30$ with $N=1000$ neurons per layer.  The asymptotic population variance $q^{\infty}$ is obtained by averaging the population variance in the last $5$ layers.  
}
\label{nonlin_dyn}
\vspace{-.2in}
%\end{wrapfigure}
\end{figure}
This map from input to output variance is numerically computed for $g=1$ and $\phi(x) = \tanh(x)$ in Fig. \ref{nonlin_dyn}, left (other values of $g$ yield a simple multiplicative scaling of this map).  This recursion relation has a stable fixed point $q^\infty(g)$ obtained by solving the nonlinear fixed point equation
\begin{equation}
q^{\infty} = g^2 \, \int \mathcal{D}z \,\phi \big(\sqrt{q^{\infty}} z \big)^2.
\label{eq:qfixed}
\end{equation}
Graphically, solving this equation corresponds to scaling the curve in Fig. \ref{nonlin_dyn} left by $g^2$ and looking for intersections with the line of unity.  For $g<1$, the only solution is $q^{\infty}=0$.  For $g>1$, this solution remains, but it is unstable under the recurrence \eqref{eq:qrec}.  Instead, for $g>1$, a new stable solution appears for some nonzero value of $q^\infty$. The entire set of stable solutions as a function of $g$ is shown as the red curve in Fig. \ref{nonlin_dyn} right.  It constitutes a theoretical prediction of the population variance at the deepest layers of a nonlinear network as the depth goes to infinity.  It matches well for example, the empirical population variance obtained from numerical simulations of nonlinear networks of depth $30$ (blue points in Fig. \ref{nonlin_dyn} right).   

Overall, these results indicate a dynamical phase transition in neural activity propagation through the nonlinear network as $g$ crosses the critical value $g_c = 1$.  When $g > 1$, activity propagates in a chaotic manner, and so $g=1$ constitutes the edge of chaos.

\end{document}